\documentclass[conference]{IEEEtran}
\IEEEoverridecommandlockouts
\usepackage{cite}
\usepackage{amsmath,amssymb,amsfonts}
\usepackage{algorithmic}
\usepackage{graphicx}
\usepackage{subcaption}
\usepackage{threeparttable}  
\usepackage{textcomp}
\usepackage{multirow} 
\usepackage{caption}
\usepackage{svg}
\usepackage{float} 
\usepackage{booktabs} 
\DeclareMathOperator{\relu}{ReLU}
\usepackage{tikz}
\usepackage{xcolor}
\def\BibTeX{{\rm B\kern-.05em{\sc i\kern-.025em b}\kern-.08em
    T\kern-.1667em\lower.7ex\hbox{E}\kern-.125emX}}
\begin{document}

\title{Unfolding ADMM for Enhanced Subspace Clustering of Hyperspectral Images\\
\thanks{This work was supported by the Research Foundation - Flanders, under the grant G094122N (project SPYDER), the Flanders AI Research Programme under Grant 174B09119, and by the China Scholarship Council under Grant 202106150007}
}

\author{
    \IEEEauthorblockN{
        Xianlu Li\IEEEauthorrefmark{1}, 
        Nicolas Nadisic\IEEEauthorrefmark{1}, 
        Shaoguang Huang\IEEEauthorrefmark{2} and
        Aleksandra  Pi\v{z}urica\IEEEauthorrefmark{1}
    }
    \IEEEauthorblockA{
        \IEEEauthorrefmark{1}Department of Telecommunications and Information Processing, Ghent University, Belgium.\\
    }
    \IEEEauthorblockA{
        \IEEEauthorrefmark{2}School of Computer Science, China University of Geosciences, Wuhan, China.\\
    }
}
\maketitle

\begin{abstract}

Deep subspace clustering methods are now prominent in clustering, typically using fully connected networks and a self-representation loss function. However, these methods often struggle with overfitting and lack interpretability. In this paper, we explore a clustering approach based on deep unfolding. By unfolding iterative optimization methods into neural networks, this approach offers enhanced interpretability and reliability compared to data-driven deep learning methods, and greater adaptability and generalization than model-based approaches. Unfolding techniques has become widely used in inverse imaging problems, such as image restoration, reconstruction, and super-resolution, but has not been sufficiently explored yet in the context of clustering. In this work, we introduce an innovative clustering architecture for hyperspectral images (HSI) by unfolding an iterative solver based on the Alternating Direction Method of Multipliers (ADMM) for sparse subspace clustering.
To our knowledge, this is the first attempt to apply unfolding ADMM for computing the self-representation matrix in subspace clustering. Moreover, our approach captures well the structural characteristics of HSI data by employing the K nearest neighbors algorithm as part of a structure preservation module. Experimental evaluation of three established HSI datasets shows clearly the potential of the unfolding approach in HSI clustering and even demonstrates superior performance compared to state-of-the-art techniques.



\end{abstract}


\section{Introduction}

    Hyperspectral images (HSI), captured by sensors that measure reflectance across numerous spectral bands, provide detailed material information of the landcover. HSI have been widely applied in fields like agriculture\cite{Agriculture}, defense\cite{Military}, and environmental monitoring\cite{Environment}, where clustering of image pixels plays an important role. Various clustering techniques such as K-means\cite{Kmeans}, C-means\cite{FCM}, spectral clustering\cite{spectral_clustering}, and Finch\cite{finch} have been developed for HSI data. However, due to the high dimensionality and variability within HSI data, traditional methods often struggle\cite{Huang2023Model}. 
    
    Subspace clustering resolves these challenges by partitioning high-dimensional data into lower-dimensional subspaces, each subspace corresponds to a distinct class. Within each subspace, data points are represented as linear combinations of each other, forming a self-representation matrix. Then a symmetric non-negative similarity matrix is obtained from the self-representation matrix. At last, spectral clustering is applied to get the cluster assignment. In subspace clustering, traditional optimizing methods such as the Alternating Direction Method of Multipliers (ADMM)\cite{boyd2011distributed} are applied to solve the optimization problem that yields the self-representation matrices. Traditional subspace clustering employs matrix representation to obtain shallow features for clustering, which have difficulties in handling non linearly separable data.
    Thus, deep clustering methods like Deep Embedded Clustering (DEC)\cite{xie2016unsupervised} and Joint Unsupervised Learning (JULE)\cite{yang2017towards} have been proposed. Nevertheless, they encounter difficulties with extensive data requirements and limited interoperability.
    
    Recently, deep subspace clustering methods that combine subspace clustering and deep learning have set the state of the art in the HSI clustering area\cite{DSCNet,cai2021hypergraph,li2023model}. However, these methods also have limitations, including a tendency to overfit and a lack of interpretability.
    
    The unfolding approach, which transforms traditional optimization iterative steps into neural network layers for more effective and adaptive problem-solving strategies, offers a novel approach compared to pure gradient descent methods.
    It has achieved impressive performance in various fields such as image denoising\cite{zeng2023degradation}, compressed sensing\cite{kouni2022admm}, hyperspectral image unmixing\cite{zhou2021admm}, etc.
    It has also been applied on dictionary learning to cluster data\cite{tankala2021clustering}. 
    
    In this paper, we introduce a novel unfolding approach for clustering HSI. This approach is built upon a deep auto-encoder for extracting spatial features and adaptive nonlinear map data into a latent space. Specifically, We unfold an ADMM-based sparse subspace clustering algorithm into a neural network architecture to obtain the self-representation matrix. 
    To maintain the intrinsic structure of the data, we employ nearest neighbors adjacency matrices to initialize the ADMM optimization process and promote uniformity in representation parameters among neighboring data points, thus preserving the data structure. The main contributions of this work are summarized as follows:
    
    \begin{itemize}
        \item We develop an unfolding network for calculating the self-representation matrix for subspace clustering.
        \item We incorporate structural priors into the optimization process and encourage similar representation parameters of neighboring data points to preserve the data structure.
        \item We evaluate our model on three HSI datasets, evidencing superior performance in comparison to current state-of-the-art methods.
    \end{itemize} 
    
    This paper is organized as follows.
    In section~\ref{sec:relwork}, we explore related research.
    We introduce our proposed approach in section~\ref{sec:proposed}. 
    Afterward, we evaluate the effectiveness of our method through experiments in section~\ref{sec:xp}, and section~\ref{sec:conclu} concludes this article.


\section{Related work}\label{sec:relwork}
\subsection{Subspace clustering}
    Subspace clustering methods typically consist of two main steps, (1) calculating the similarity matrix and (2) applying spectral clustering on the similarity matrix to get the clustering result. To obtain the similarity matrix, most state-of-the-art methods apply a self-representation model with a sparse regularization as follows:
    \begin{equation}
        \min||\mathbf{X}-\mathbf{XC}||^{2}_{F} + \lambda||\mathbf{C}||_1 \quad \text{s.t.} \quad diag(\mathbf{C})=0,
    \end{equation}
    where $\mathbf{X} \in \mathbb{R}^{d\times n}$ is the data matrix, and $\mathbf{C} \in \mathbb{R}^{n\times n}$ is the self-representation matrix. The constraint $diag(\mathbf{C})=0$ prevents a trivial solution where each data point is represented by itself. To solve this optimization problem, algorithms such as ADMM are often applied. 
    Once we have the self-representation matrix, we can compute a symmetric and non-negative similarity matrix as follows:
    \begin{equation}
        \mathbf{S} = \frac{1}{2}(|\mathbf{C}|+|\mathbf{C}|^T).
    \end{equation}
    
    The similarity matrix contains the pairwise affinity between data points, and finally, spectral clustering is applied to get the cluster label of every data point.

\subsection{Data-driven deep clustering}
    In data-driven deep clustering, deep neural networks are employed to directly learn representations of data. These approaches can be divided into two types: the first type applies traditional clustering algorithms on features extracted by the neural network \cite{DEN, PARTY}; the second involves a joint optimization of the clustering network and feature extraction process \cite{xie2016unsupervised, yang2017towards}.
    
\subsection{Deep subspace clustering}
    Traditional subspace clustering assumes that high-dimensional data points are sampled from a union of low-dimensional subspaces. Deep subspace clustering\cite{DSCNet,li2023model} embed this assumption in deep learning. Unlike traditional subspace clustering methods that apply an iterative optimization method for calculating the self-representation matrix, some deep subspace clustering methods employ fully connected networks, where the parameters of the neural network serve as self-representation variables. In this approach, a self-representation loss function is applied, enabling the derivation of self-representation matrices through gradient descent\cite{DSCNet,li2021self}. Some recent methods learn the subspace basis\cite{li2023model} to enable more efficient clustering in large datasets. 
    
\subsection{Deep unfolding}
    Deep unfolding \cite{monga2021algorithm} merges traditional optimization with deep learning, transforming iterative optimization steps into neural network layers, where each layer mirrors a step in the optimization algorithm, this enables effective variable updates and enhances the interpretability of the neural network.

    Unfolding techniques have been applied in various inverse problems in imaging like image denoising\cite{zeng2023degradation}, compressive sensing\cite{kouni2022admm}, and hyperspectral image unmixing\cite{zhou2021admm}. These problems focus on recovering data from degraded or incomplete inputs. In the clustering task, the goal is to uncover the underlying data structure. In \cite{tankala2021clustering}, the authors apply deep dictionary learning to clustering by unfolding an iterative soft thresholding algorithm (ISTA). This approach is quite different from self-representation learning, likely to cause overfitting and loss of data structure. In traditional sparse subspace clustering, the self-representation matrix can capture the data structure well. However, obtaining this matrix effectively with unfolding techniques remains a challenge. 

\section{Proposed method}\label{sec:proposed}
\begin{figure*}
    \centering
    \includegraphics[width=0.7\textwidth]{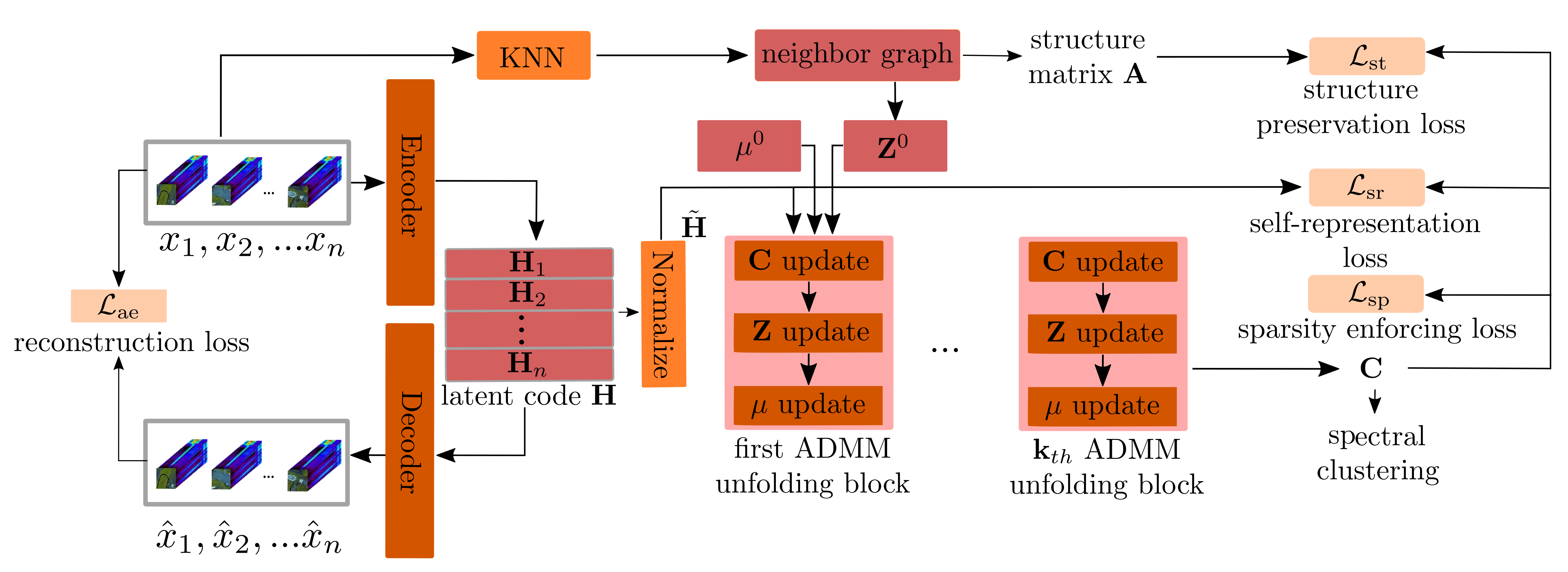}
    \caption{The structure of the proposed method. As shown in the figure, initially, the HSI patches {$x_1,x_2,...x_n$} are fed into a deep auto-encoder to extract latent representations $\mathbf{H}$, with a reconstruction loss. This latent representation is then transposed and normalized, resulting in $\Tilde{\mathbf{H}}$, which is fed into an ADMM unfolding network to produce the self-representation matrix $\mathbf{C}$. To improve the quality of $\mathbf{C}$, the loss of self-representation, the loss of structure-preservation, and the sparsity loss are applied.}

    \label{main structure}
\end{figure*}

Here, we present our proposed methodology, which encompasses three distinct parts: (1) the design of the ADMM unfolding network of self-representation model for subspace clustering; (2) feature learning with ADMM unfolding for enhanced self-representation; (3) structure preservation in self-representation. The architecture of our model is depicted in Figure \ref{main structure}. 

\subsection{Unfolding ADMM optimization for self-representation}
    In this section, we detail the process of unfolding the ADMM algorithm for solving a self-representation model in subspace clustering. As far as we are aware, this marks the first instance of applying the unfolding approach to obtain a self-representation matrix for clustering purposes. To effectively leverage the learning capabilities of neural networks, we have restructured the original optimization problem of the self-representation model as follows:
        \begin{equation}
            \min||\mathbf{X}-\mathbf{YC}||^2_{F} + \lambda||\mathbf{C}||_{1} \quad \text{s.t.} \quad diag(\mathbf{C})=0,
        \end{equation}
      Here, $\mathbf{Y}$ is a learnable auxiliary variable, initially set to $\mathbf{X}$ and highly correlated with it. Setting $diag(\mathbf{C})=0$ prevents data points from representing themselves, which is a common constraint in self-representation models.
    Similarly, according to the principle of the ADMM algorithm, we can create an augmented Lagrangian function:
    \begin{equation}
        L = ||\mathbf{X}-\mathbf{YC}||^2_{F} + \lambda||\mathbf{Z}||_{1} + \langle \mu, \mathbf{C}-\mathbf{Z} \rangle + \frac{\rho}{2}||\mathbf{C}-\mathbf{Z}||^2_{F},
    \end{equation}
    where $\mathbf{Z}\in \mathbb{R}^{n\times n}$ is the auxiliary variable, introduced to decompose the optimization of $\mathbf{C}$ into separate optimizations for $\mathbf{C}$ and $\mathbf{Z}$. $\mu$ and $\rho$ are used for enforcing constraints and controlling the convergence of the optimization problem. Then we will optimize the variables one by one as follows:\\
    \textbf{Update} $\mathbf{C}$:\\
    \begin{equation}
        \mathbf{C^{k+1}} = (2\mathbf{Y}^T\mathbf{Y}+\rho)^{-1}(2\mathbf{Y}^T)\mathbf{X} - (2\mathbf{Y}^T\mathbf{Y}+\rho)^{-1}(\mu^{k} -\rho\mathbf{Z^{k}}).
    \end{equation}
    When integrating this into a fully connected neural network, similar to\cite{zhou2021admm} the following expressions emerge: \\
    \begin{equation}
    \begin{aligned}
        &\mathbf{C^{k+1}} = \mathbf{W}\mathbf{X}-\mathbf{B}(\mu^{k} -\rho\mathbf{Z^{k}}) \\
        &\mathbf{W} = (2\mathbf{Y}^T\mathbf{Y}+\rho)^{-1}(2\mathbf{Y}^T) \\ 
        &\mathbf{B} = (2\mathbf{Y}^T\mathbf{Y}+\rho)^{-1},
    \end{aligned}
    \end{equation}
    where $\mathbf{W}$ and $\mathbf{B}$ are the learnable parameter of fully connected layers, $\rho$ is also a learnable parameter for better updating.\\
    \textbf{Update} $\mathbf{Z}$:\\
    \begin{equation}
    \mathbf{Z}^{k+1} = \text{SoftThresholding}\left(\mathbf{C}^{k+1} + \frac{\mu^k}{\rho}, \frac{\lambda}{\rho}\right) ,
    \end{equation}
    where the SoftThresholding is defined as follows:
    \begin{align}
        \text{SoftThresholding}(x, \theta) = 
            \begin{cases} 
            x - \theta & \text{if } x > \theta \\
            0 & \text{if } |x| \leq \theta \\
            x + \theta & \text{if } x < -\theta 
            \end{cases}.
    \end{align}
    In the unfolding network, we utilize a rectified linear unit (ReLU) activation function. Instead of using $\frac{\lambda}{\rho}$, here we apply a learnable threshold initialized with the value of 0.005. This choice is motivated by the effectiveness of ReLU in facilitating gradient propagation. 
    \begin{equation}
         \mathbf{Z^{k+1}} = \relu(|\mathbf{C}^{k+1} + \frac{\mu^{k}}{\rho}| - \text{threshold})\cdot \text{sgn}(\mathbf{C}^{k+1} + \frac{\mu^{k}}{\rho}),
    \end{equation}
    where $\text{sgn}()$ is the sign function.
    To prevent trivial solution, we set diagonal of $\mathbf{Z^{k+1}}$ to zero:
    \begin{equation}
        diag(\mathbf{Z^{k+1}})=0 .
    \end{equation}
    \textbf{Update} $\mathbf{\mu}$:\\
    \begin{equation}
        \mathbf{\mu}^{k+1} = \mathbf{\mu}^{k} + \rho\cdot(\mathbf{C^{k+1}}-\mathbf{Z^{k+1}}).
    \end{equation}
    After completing the optimization iterations, we obtain the self-representation matrix $\mathbf{C}$ from the ADMM network. To ensure the high representation quality of the self-representation matrix $\mathbf{C}$ obtained from the ADMM network, we aim to ensure that $\mathbf{C}$, derived from the auxiliary variables $\mathbf{Y}$, can accurately reconstruct $\mathbf{X}$, as demonstrated in the following equation:
    \begin{equation}
        \mathcal{L}_{\text{sr}} = \frac{1}{n} \sum_{i=1}^{n} \left\| \mathbf{X}_{i} - \mathbf{XC}_{i} \right\|_{2},
        \label{recons}
    \end{equation}
    where $\mathbf{X}_{i}$ and $\mathbf{XC}_{i}$ is the $i_{th}$ column of $\mathbf{X}$ and $\mathbf{XC}$, correspond to $i_{th}$ data point and its reconstruction. 
    To encourage sparsity by neural network optimization, an $l_1$ regularization was also applied shown as follows:
    \begin{equation}
        \mathcal{L}_{\text{sp}} = \frac{1}{n}||\mathbf{C}||_{1} .
    \end{equation} 
    In this way, we unfold the ADMM solver of the self-representation model for subspace clustering into a neural network.

\subsection{Auto-encoder with unfolding ADMM}
    To leverage the spatial information in HSI data and enhance the handling of nonlinear features, we apply a convolutional auto-encoder to jointly optimize with the unfolding network. 
    First, to ensure the latent representation retains substantial information from original data, the mean square error loss is applied:
    \begin{equation}
        \mathcal{L}_{\text{ae}} = \frac{1}{n} \sum_{i=1}^{n} ||{x}_i - \hat{x}_i||_F^2,
    \end{equation}
    where ${x}_i$ and $\hat{x}_i$ are the $i_{th}$ HSI patch and its reconstruction value. Then the unfolding ADMM is applied to this latent code, and we redefine $\mathcal{L}_{\text{sr}}$ from \eqref{recons} using the latent code:
    \begin{equation}
    \mathcal{L}_{\text{sr}} = \frac{1}{n} \sum_{i=1}^{n} \left\| \Tilde{\mathbf{H}}_{i} - \mathbf{\Tilde{H}C}_{i} \right\|_{2},
    \end{equation}
    where $\Tilde{\mathbf{H}} \in \mathbb{R}^{l\times n}$ is the normalized transpose of the latent code matrix of original data, and $l$ is the dimension of latent code, $\mathbf{C}_{i}$ is the $i_{th}$ column of $\mathbf{C}$.

\subsection{Structure preservation module}
    In our approach, we exploit the K-nearest neighbors (KNN) algorithm to capture the structural characteristics of HSI data, resulting in two distinct adjacency matrices. The first matrix, encompassing a larger set of neighbors, is utilized for initializing the matrix $\mathbf{Z}$. The second matrix, with a smaller neighbor set, aims to ensure consistency in the self-representation matrix, particularly by maintaining similar representation characteristics among neighboring data points. This is important for preserving the structural integrity of the dataset. Formally, we express the structure preservation loss as follows:
    \begin{equation}
    \mathcal{L}_{\text{st}} = \sum_{i,j} \mathbf{A}_{ij}||\mathbf{C}_{i}-\mathbf{C}_j||^2 = \mathbf{Tr}(\mathbf{C}^T\mathbf{L}\mathbf{C}).
    \end{equation}
    Here, $\mathbf{A}$ represents the adjacency matrix, where $\mathbf{A}_{ij}=1$ indicates that the $i_{th}$ and $j_{th}$ data points are neighbors. $\mathbf{L}$ is the Laplacian matrix derived from $\mathbf{A}$, and $\mathbf{Tr}$ refers to the trace calculation. $\mathbf{C}_{i}$ corresponds to the $i_{th}$ column of the self-representation matrix. 

\subsection{Model Training}

The training process of our model is structured into two primary steps for effective learning and we will use all data in both steps:

\begin{enumerate}
    \item \textbf{Pretraining of deep autoencoder:} Initially, the deep autoencoder is pretrained. During this phase, the network is trained in an unsupervised manner with clustering data, which allows it to learn useful features and patterns. This phase focuses on minimizing the reconstruction loss, expressed as $\mathcal{L}_{\text{ae}}$.

    \item \textbf{Training the complete network:} After pretraining, the entire network start training. This stage involves the integration of various loss components, formulated as:
    \begin{equation}
        \mathcal{L}_{\text{all}} = \mathcal{L}_{\text{ae}} + \alpha \mathcal{L}_{\text{sr}} + \beta \mathcal{L}_{\text{sp}} + \gamma \mathcal{L}_{\text{st}} .
    \end{equation}
    Here, $\alpha$, $\beta$ and $\gamma$ represent the weights assigned to $\mathcal{L}_{\text{sr}}$ (self representation loss), $\mathcal{L}_{\text{sp}}$ (sparse loss) and $\mathcal{L}_{\text{st}}$ (structure preservation loss), respectively.
\end{enumerate}
    After training process, the self-representation matrix is obtained, the spectal clustering is then applied to get the final clustering result.

\section{Experiments and results}\label{sec:xp}
    \subsection{Dataset introduction}
    Our model is evaluated using three well-known datasets: Pavia University, Salinas, and Indian Pines. 

            
            \begin{table}[!ht]
                \centering
                \begin{tabular}{@{}ccccc@{}}
                \toprule
                Dataset & Data Size & Train Samples & Number Classes & Patch Size \\
                \midrule
                Salinas & $83\times86$ & 5348 & 6 & $7\times7$ \\
                Indian\_pines & $85\times70$ & 4391 & 4 & $7\times7$ \\
                PaviaU & $100\times200$ & 6445 & 8 & $13\times13$ \\
                \bottomrule
                \end{tabular}
                \caption{The training dataset setting}
                \label{dataset}
            \end{table}
    \subsection{Experiment setting and training strategy}
    To ensure a good initial state for the optimization process, we initialize $\mathbf{Z}$ using the 30 nearest neighbors, taking advantage of the structural prior. For $\mathbf{C}$, aiming for higher accuracy, we incorporate stricter structural information by selecting the 10 nearest neighbors. For the Salinas dataset, we use 2 ADMM iterations and a small $\rho$. For the Indian Pines and PaviaU dataset, due to more complex features, we use 3 ADMM iterations and larger $\rho$ with larger update step size. 
    We employ three classical metrics in the HSI clustering domain: Accuracy (ACC), Normalized Mutual Information (NMI), and Kappa. Higher values in these metrics correspond to better performance. We provide our code and test scripts in an online repository\footnote{https://github.com/lxlscut/Unfolding-ADMM-for-Enhanced-Subspace-Clustering-of-Hyperspectral-Images}.
    The inner structures of different datasets vary, leading to differences in the weights of various components within the loss function. The specific weights and initial $\rho$ values for each dataset are presented in Table \ref{parameter}.
    \begin{table}[!ht]
    \centering
    \begin{tabular}{@{}ccccc@{}}
    \toprule
    Dataset & $\rho$ & $\alpha$ & $\beta$ & $\gamma$  \\
    \midrule
    Salinas & 0.1 & 40 & 0.1 & 0.0001\\
    Indian Pines & 0.9 & 40 & 0.3 & 0.0003\\
     PaviaU & 0.5 & 40 & 1.3 & 0.01 \\
    \bottomrule
    \end{tabular}
    \caption{The hyper-parameter setting}
    \label{parameter}
    \end{table}
    \subsection{Clustering result}
\begin{figure*}[!ht]
        \centering
        \newcommand{\insertsubfigure}[2]{
            \begin{subfigure}{0.08\textwidth} 
                \centering
                \begin{tikzpicture}
                    \node[anchor=south west,inner sep=0] (image) at (0,0) {\includegraphics[width=\linewidth]{photo/#1}};
                    \begin{scope}[x={(image.south east)},y={(image.north west)}]
                \draw[red, thick] (0.5,0.0) rectangle (1.0,0.4);
                \end{scope}
                \end{tikzpicture}
                \caption{} 
                \label{#2}
            \end{subfigure}
        }
        \newcommand{\insertsubfigureNoFrame}[2]{
            \begin{subfigure}{0.08\textwidth} 
                \centering
                \includegraphics[width=\linewidth]{photo/#1}
                \caption*{} 
                \label{#2}
            \end{subfigure}
        }
        \insertsubfigure{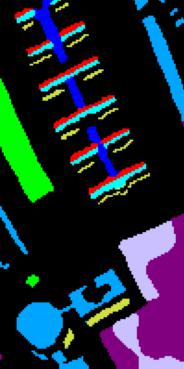}{fig:sub1}
        \insertsubfigure{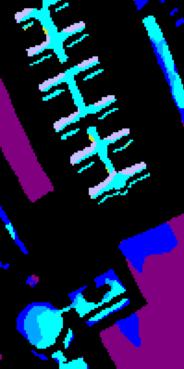}{FCM}
        \insertsubfigure{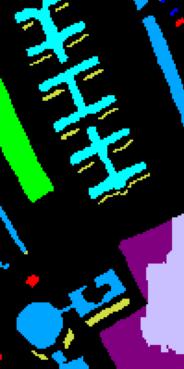}{Finch}
        \insertsubfigure{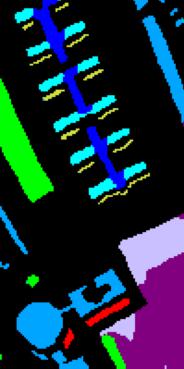}{SC}
        \insertsubfigure{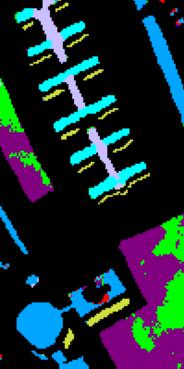}{SpectralNET}
        \insertsubfigure{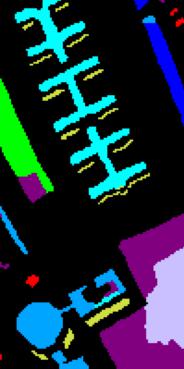}{DscNet}
        \insertsubfigure{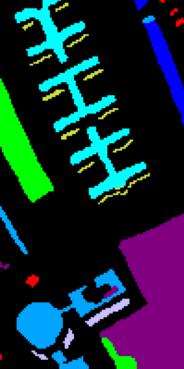}{HyperAE}
        \insertsubfigure{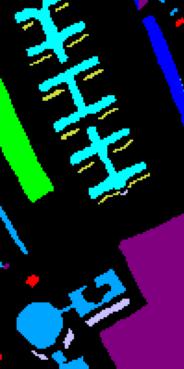}{Ours}
        \insertsubfigure{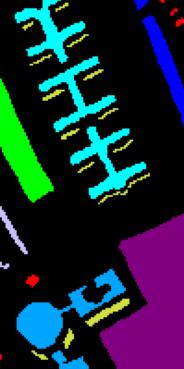}{Ground_truth}
        \insertsubfigureNoFrame{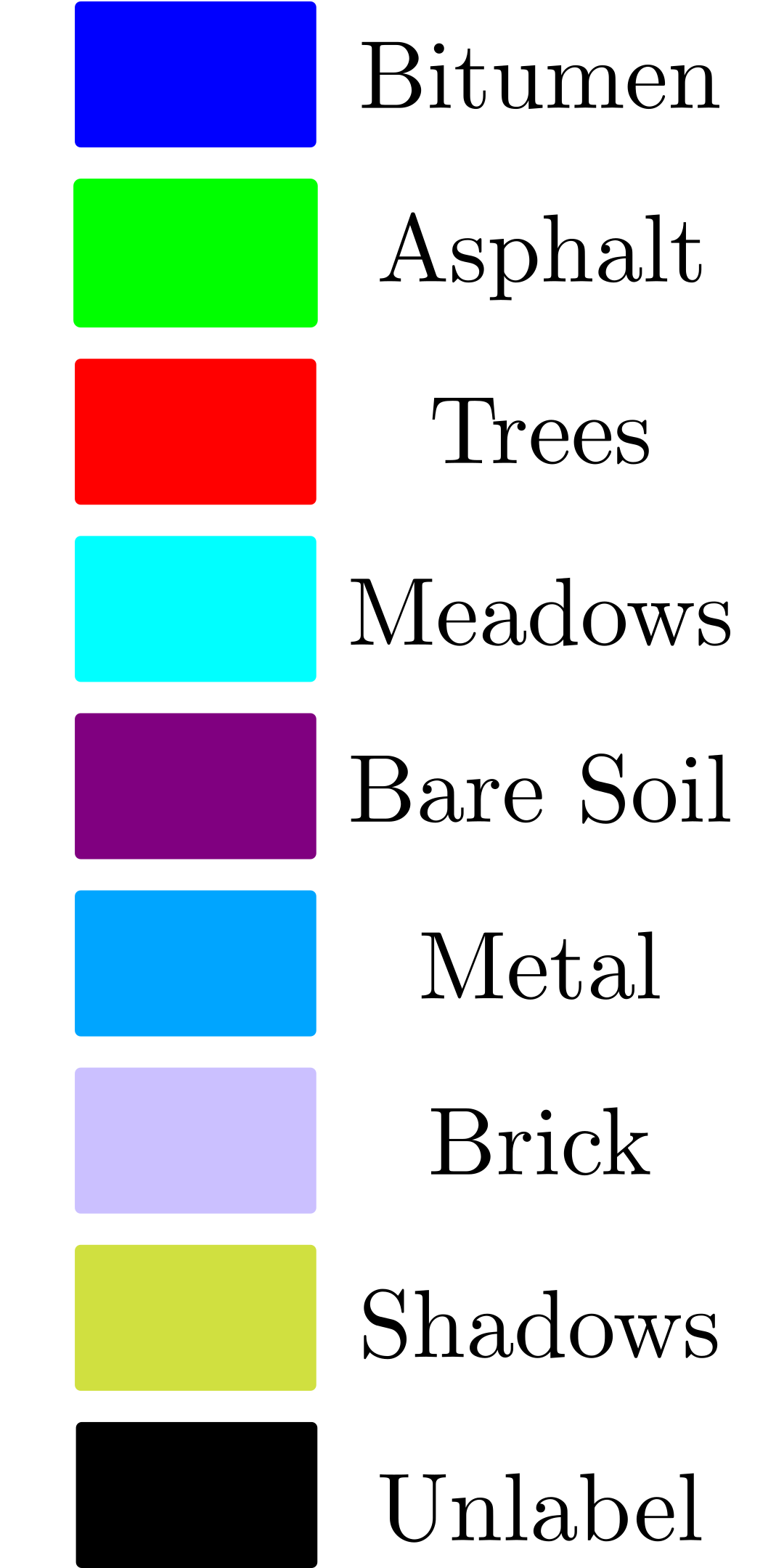}{label}
        \caption{\footnotesize Results on PaviaU: (a) Kmeans, (b) FCM, (c) Finch, (d) SC, (e) SpectralNET, (f) DscNet, (g) HyperAE, (h) Ours, (i) Ground truth}
        \label{fig:visual}
    \end{figure*}
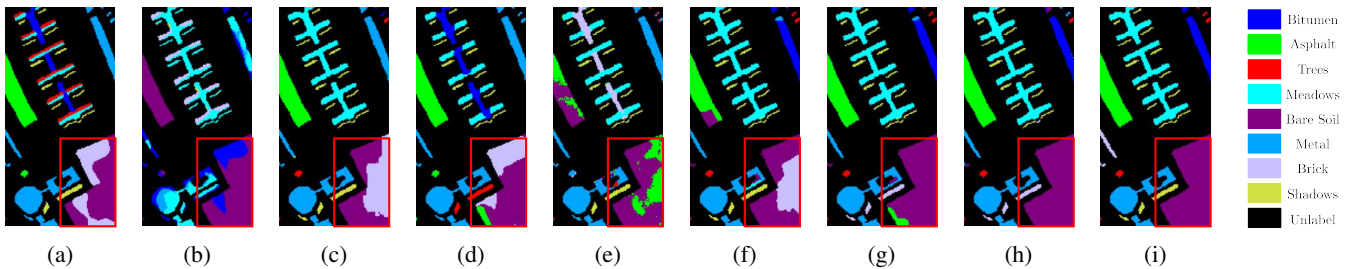

      Our model was evaluated using the described datasets, and compared with several mainstream methods. Comparative and visual results are in table \ref{comparison} and figure \ref{fig:visual}, respectively.
      \begin{table*}[!ht]
        \centering
        \begin{threeparttable}
            \begin{tabular}{@{}cccccccccc@{}}
            \toprule
            Datasets & Metrics & K-means\cite{Kmeans} & C-means\cite{FCM} & Finch\cite{finch} & SC\cite{spectral_clustering} & SpectralNet\cite{SpectralNet} & Dscnet\cite{DSCNet} & HyperAE\cite{cai2021hypergraph} & Ours \\
            \midrule
            \multirow{3}{*}{Salinas} & ACC & 0.7418 & 0.7693 & 0.7186 & 0.8037 & 0.7792& 0.9302 & \textbf{1} & \textbf{1}\\
                                      & NMI & 0.8394 & 0.7418 & 0.8141 & 0.8899 & 0.7113
                                      & 0.9081& \textbf{1} & \textbf{1}\\
                                      & KAPPA & 0.6798 & 0.7029 & 0.6554 & 0.7560 &0.7153
                                      & 0.9134 & \textbf{1} & \textbf{1}\\
            \midrule
            \multirow{3}{*}{Indian\_pines}  & ACC & 0.5994 & 0.5083 & 0.6550 & 0.6894 & 0.5655 & 0.5867&0.6557&\textbf{0.9132}\\
                                        & NMI & 0.4468 & 0.2144 & 0.5063 & 0.5754 & 0.4407& 0.4473&0.5497&\textbf{0.7951}\\
                                        & KAPPA & 0.4530 & 0.3425 & 0.5441 & 0.5101 & 0.4246 &0.4161 &0.4987&\textbf{0.8742}\\
            \midrule
            \multirow{3}{*}{PaviaU}  & ACC & 0.6239 & 0.4852 & 0.7330 & 0.6448 & 0.6282& 0.8083 & 0.9357 & $\mathbf{0.9561}$\\
                                         & NMI & 0.7858 & 0.5028 & 0.8642 & 0.7713 & 0.6684& 0.8313 & 0.9082 & $\mathbf{0.9552}$\\
                                        & KAPPA & 0.5481 & 0.3168 & 0.6764 & 0.5678 &0.5229 & 0.7616 & 0.9159 & $\mathbf{0.9421}$\\
            \bottomrule
            \end{tabular}
            \label{result}
        \end{threeparttable}
        \caption{Results of the experiments on the clustering of real-world hyperspectral images. The best results are shown in \textbf{bold}.}
        \label{comparison}
    \end{table*}
    
      From the clustering results, we can see that our method outperforms other mainstream approaches. Notably, Dscnet and HyperAE show significantly better performance compared to others. Meanwhile, graph-based methods like spectral clustering and the hierarchical clustering approach like Finch deliver much better results than centroid-based methods like K-means and C-means. It’s also important to mention that SpectralNet didn’t perform as expected, possibly due to the challenges presented by smaller datasets.
     
\section{Conclusion}\label{sec:conclu}
    We introduced a neural network built from the unfolding of ADMM and combined with an auto-encoder to derive a self-representation matrix. To the best of our knowledge, this is the first attempt to utilize unfolding ADMM for computing self-representation matrices in subspace clustering. Additionally, to preserve the inherent data structure within the self-representation matrix, we incorporate the K nearest neighbors algorithm in the structure preservation module. This regularization promotes capturing the manifold structure of HSI in the feature domain. Our methodology is evaluated on three widely used datasets, demonstrating its superior performance compared to other mainstream methods. This reflects that the unfolding technology has advantages over traditional deep subspace clustering in solving self-representation-based subspace clustering problems.
\bibliographystyle{IEEEtran}
\bibliography{conference_101719}

\end{document}